\documentclass{article}

     \usepackage[final]{neurips_2019}

\usepackage[utf8]{inputenc} 
\usepackage[T1]{fontenc}    
\usepackage{hyperref}       
\usepackage{url}            
\usepackage{booktabs}       
\usepackage{amsfonts}       
\usepackage{nicefrac}       
\usepackage{microtype}      
\usepackage{natbib}
\usepackage{amsmath,amsfonts,amssymb}
\usepackage{scalerel}
\usepackage{multirow}
\usepackage{marvosym}
\usepackage{wrapfig}
\usepackage{wrapfig}

\usepackage{amsthm}
\usepackage{amsmath}
\usepackage{amssymb}
\usepackage{ upgreek }
\usepackage{dsfont}
\usepackage{mathrsfs}
\usepackage{graphicx}
\usepackage{wrapfig}

\usepackage{tikz-cd}

 \newcommand*{\crosssymbol}{%
    \text{%
      \raisebox{1ex}{%
        \makebox[0pt][l]{%
          \rule[-.2pt]{.75ex}{.4pt}%
        }%
        \makebox[.75ex]{%
          \rule[-1ex]{.4pt}{1.5ex}%
        }%
      }%
    }%
}  
\newcommand*\samethanks[1][\value{footnote}]{\footnotemark[#1]}

\usepackage{graphicx}
\usepackage{wrapfig}
\usepackage{array}
\usepackage{tikz-cd}
\title{Graph Attentional Autoencoder for Anticancer Hyperfood Prediction}

\author{%
  \begin{tabular}{ c c c }
  Guadalupe Gonzalez$^{1}\thanks{Equal contribution.}$ & Shunwang Gong$^{1}$\samethanks
  & Ivan Laponogov$^{1}$\\[3mm] 
  Kirill Veselkov$^{1}$
  & Michael Bronstein$^{1,2,3}$
 ~~~
  \end{tabular}
  \vspace{4mm}\\
  \begin{tabular}{ c c c }
  $^1$Imperial College London& $^2$Twitter & $^3$USI Lugano  \\ 
 United Kingdom & United Kingdom & Switzerland
  \end{tabular}
  }
\begin{document}

\maketitle

\begin{abstract}
Recent research efforts have shown the possibility to discover anticancer drug-like molecules in food from their effect on protein-protein interaction networks, opening a potential pathway to disease-beating diet design. We formulate this task as a graph classification problem on which graph neural networks (GNNs) have achieved state-of-the-art results. However, GNNs are difficult to train on sparse low-dimensional features according to our empirical evidence. Here, we present graph augmented features, integrating graph structural information and raw node attributes with varying ratios, to ease the training of networks. We further introduce a novel neural network architecture on graphs, the \textit{Graph Attentional Autoencoder} (GAA) 
to predict food compounds with anticancer properties based on perturbed protein networks. We demonstrate that the method outperforms the baseline approach and state-of-the-art graph classification models in this task. 

\end{abstract}


\section{Introduction}



    Even though cancer is the second leading cause of death in the United States~[\citenum{Siegel2019}], almost 40\% of all cancers are preventable through dietary and lifestyle changes. It has been experimentally shown that fruits and vegetables are particularly rich in various types of cancer-beating molecules ~[\citenum{Research1997, Donaldson2004}].  In light of this, increasing research efforts aim at elucidating the biological compounds and molecular mechanisms responsible for the observed anticancer properties of foods.
    
	Experimental studies have elucidated disease-preventing properties of several food compounds. However, the low efficiency and high cost of experimental studies have motivated the adoption of computational techniques to study food compounds' properties. \citet{Veselkov2019HyperFoods:Foods} introduced a novel study on large-scale prediction of anticancer food compounds based on their commonality to act on protein-protein networks similarly to clinically-approved anticancer drugs. Using Random Walk with Restarts (RWR) to generate vector representations of protein-protein networks characterizing drugs, a supervised method was used to classify drugs into anticancer and non-anticancer classes. The trained classifier was then used to predict anticancer food compounds. We formulate this problem as a graph classification task in which all graphs (drugs and food compounds) share the same topology (human protein-protein network) differing only in node features (protein targets of drugs: binary feature on each node).
	
    \textbf{Graph neural networks.}  Graph neural networks, developed as part of the recent trend of deep learning on graphs [\citenum{Bronstein2017GeometricData, Hamilton2017}],  have been particularly useful for classifying properties of graphs corresponding to different molecules ~[\citenum{Duvenaud, Gilmer2017NeuralChemistry, Xu2019HowNetworks}]. These graph classification models have produced state-of-the-art results in graph classification benchmarks [\citenum{Verma2019GraphNetworks, Xu2019HowNetworks, Yanardag2015DeepKernels}]. However, one remarkable difference between drug-protein datasets and the common benchmarks is that, in the former, only sparse $1$-dimensional binary node features (1.36\% positive ratio) are provided, and drugs (or food compounds) share the same topology, which makes it a challenge for the majority of graph classification models. 
    
    \textbf{Contribution.} To address the difficulty of training GNNs with sparse $1$-dimensional node features, we propose to convert raw features to the continuous space by integrating the structural information of the graph and raw node attributes, with varying ratios. The model proposed here, \textit{GAA}, is capable of learning from these graph augmented features and finding the balanced integration ratio, outperforming baseline models and state-of-the-art graph classification models in this task. Additionally, we show that we are able to inject prior biological knowledge in the neural network architecture producing biologically-meaningful graph embeddings.
    %


\section{Our Approach} \label{sec:approach}


In this section, we introduce our model, \textit{GAA}. The model consists of three major steps: (i) it converts raw discrete node feature vectors to the continuous space by integrating them with the graph topology; (ii) a graph attentional autoencoder further fuses the structural information of the graph and its node attributes, generating low-dimensional graph embeddings in the middle layer; (iii) embeddings are then fed into multilayer perceptrons for prediction.

\textbf{Definitions.} We consider an undirected graph $\mathcal{G}=(\mathcal{V}, \mathcal{E})$ with its node feature matrix $X \in \mathbb{R}^{N \times F}$ and adjacency matrix $A$, where $N$ is the number of nodes, and $F$ is the number of features in each node. 

\textbf{Graph augmented features.} To address the issue of low dimensional discrete node features, we consider enhancing node feature vectors by integrating raw features and the graph topology. Inspired by the RWR algorithm~[\citenum{PanAutomatic290}], we extend its definition to allow diffusion of node attributes as $X^{t + 1}  = \alpha X  + (1 - \alpha) \hat{A}  X^{t}$, where $\hat{A}$ is the column-normalized adjacency matrix, and $\alpha$ denotes the restart probability controlling the trade-off between prior information and graph smoothing.  $\alpha \to 1$ represents higher weight on node prior information (or attribute), whereas $\alpha \to 0$ represents higher weight in diffusing node attributes to the whole graph. This process converges to a steady-state distribution:
\begin{align} 
    X^{\ast}  = \alpha (I - (1 - \alpha) \hat{A})^{-1} X= k_{\alpha} X \label{eq:rwr}      
\end{align}
\noindent where $X^{\ast} \in \mathbb{R}^{N \times F}$ is the steady-state node feature matrix and $x^{\ast}_{ij}$ denotes the normalized proximity (or importance) score of node $i$ with respect to nodes with nonzero initial $j$-th feature. Based on Eq. \ref{eq:rwr}, we define the graph-augmented features as follows:
\begin{equation}
    X^{\mathcal{G}} = \left[ k_{\alpha_1}X, k_{\alpha_2}X, ..., k_{\alpha_t}X \right] \in \mathbb{R}^{N \times F'}
\end{equation}
\noindent with $F' = tF$ and where $[k_{\alpha_1}X, k_{\alpha_2}X, ..., k_{\alpha_t}X]$ are the concatenated feature matrices with varying ratios of prior information and network smoothing integration. The graph attentional autoencoder is capable of learning from $X^{\mathcal{G}}$ to find the optimal diffusion ratio.  

\textbf{Graph attentional encoder.} To allow the network propagation to leverage the unique property of our graph augmented features, our encoder is designed by stacking two layers of graph attentional networks (GATs) [\citenum{Velickovi2018GraphNetworks}] in the first two layers, which allows for (implicitly) assigning different importances to different nodes within a neighborhood. The output feature matrix of the GAT layer is described as:
\begin{align}
    H^{\ast} = \sigma \left( \mathbin\Vert_{\substack{k = 1}}^K \sum_{j \in \mathcal{N}_i} \alpha_{ij}^k W^k H  \right)
\end{align}
\noindent where $\mathbin{\Vert}$ represents concatenation and $\sigma$ is a nonlinear activation function, and the input of the first GAT layer, $H^{(0)} \in \mathbb{R}^{F' \times N}$ denotes the transpose of the graph augmented feature matrix $X^{\mathcal{G}}$.  The attentional coefficients $\alpha_{ij}^k$ are computed as:
\begin{align}
        \alpha_{ij} =
        \frac{
        \exp\left(\mathrm{LeakyReLU}\left(\vec{a}^{\top}
        [W\vec{h}_i \, \Vert \, W\vec{h}_j]
        \right)\right)}
        {\sum_{k \in \mathcal{N}(i)}
        \exp\left(\mathrm{LeakyReLU}\left(\vec{a}^{\top}
        [W\vec{h}_i \, \Vert \, W\vec{h}_j]
        \right)\right)}.
\end{align}
\noindent where $\cdot ^ {\top}$ represents transposition and $\mathbin\Vert$ is the concatenation operation. $\vec{\alpha} \in \mathbb{R}^{2F'}$ and $W \in \mathbb{R}^ {F'' \times F'}$ are weight parameters. Note that to stabilize the model training, we do not use multi-heads on the second GAT layer. 

To convert each graph to an finite dimensional vector in $\mathbb{R}^D$, we define a pooling layer based on our prior knowledge of the graph, which follows the previous GAT layers. Suppose we know any node $i$ belongs to a finite set of supermodules $U_j = \{ y_1, y_2, ..., y_k \}$ with $ D = |\cup {U_j}| < |\mathcal{V}|$, and each supermodule $y_i$ contains a finite set of nodes $\{ i_1, i_2, ..., i_t \}$, we aggregate node feature vectors $\{ \vec{x}_{i_1}, \vec{x}_{i_2}, ..., \vec{x}_{i_t} \}$ with a differentiable, permutation invariant function, \textit{e.g.}, sum, mean or max. We define it as \textit{SupPool}, namely, supermodule pooling: 
\begin{align}
    \vec{x}_{y_i} = \mathrm{SupPool}_{i \in y_i} (\vec{x}_i)
\end{align}
\noindent which yields the output node feature matrix $X_{y} \in \mathbb{R}^{D \times  C}$ assume the input feature dimensionality is $C$. The embedding vector $\vec{z} \in \mathbb{R}^D$ is then obtained from $\vec{z}^{\top} = \vec{w}^{\top} X^{\top}_y + \vec{b}^{\top}$
\noindent where $\vec{w} \in \mathbb{R}^{C}$ and $\vec{b} \in \mathbb{R}^{D}$ are weight and bias vector respectively. 

\textbf{Graph attentional decoder.} Our decoder is used to recover the graph augmented features $X^{\mathcal{G}}$. We calculate the reconstructed feature matrix $\hat{X}^{\mathcal{G}}$ as follows:
\begin{align}
    \hat{X}^{\mathcal{G}} = \sum_{j \in \mathcal{N}_i} \alpha_{ij} W^{\crosssymbol}_2 \, \mathrm{GAT} (W^{\crosssymbol}_1 \vec{z}+\vec{b}, A),  \, \mathrm{with} \, \vec{z} = \mathrm{Enc} (X^{\mathcal{G}}, A).
\end{align}
\noindent where $W^{\crosssymbol}_1 \in \mathbb{R}^{|\mathcal{V}| \times D}$ and $W^{\crosssymbol}_2 \in \mathbb{R}^{F' \times F}$ are weight matrices, assuming the dimensionality of the output feature matrix of GAT layer is $F$.   

\textbf{Multilayer perceptrons.} Given a set of graph embeddings $\{ \vec{z}_{\mathcal{G}_1}, \vec{z}_{\mathcal{G}_2}, ..., \vec{z}_{\mathcal{G}_k} \}$ and the corresponding graph labels $\{ y_1, ...., y_k \}$, the aim of the MLP is to learn a mapping function $f: \vec{z}_{\mathcal{G}_k} \to y_k$. The output of the MLP is obtained from the below equation:
\begin{align}
    \vec{y} = \sigma_2 \left( W_2 \, \sigma_1 \left( W_1 \vec{z} + \vec{b}_1 \right) + \vec{b}_2 \right )
\end{align}
\noindent where $\sigma_1$ and $\sigma_2$ are the ELU and softmax function respectively. $W_1, W_2$ and $\vec{b}_1, \vec{b}_2$ are weight and bias matrices respectively.

\textbf{Optimization.} The model loss combines the reconstruction loss $\mathcal{L}_r = || \mathrm{Enc} \circ \mathrm{Dec} \, (X^{\mathcal{G}}, A) - X^{\mathcal{G}} ||_2$ encouraging the encoder-decoder pair to be a nearly identity transformation, and the cross-entropy loss $\mathcal{L}_c$. We optimize the total loss computed as $\mathcal{L} = \mathcal{L}_c + \gamma \mathcal{L}_r $. $\gamma$ can be considered as an equilibrium term $\gamma \in [0, 1]$, which is used to maintain the balance of the loss expectation of the graph attentional autoencoder and the MLP.


\section{Evaluation and Discussion} \label{sec:eva}

\textbf{Datasets.} We followed the procedure similar to \citet{Veselkov2019HyperFoods:Foods} to compile our datasets: a human protein-protein network, drug and food compounds features. We filtered the human protein-protein network to include only experimentally-validated interactions, removed isolated nodes and kept the biggest connected component of the network (15,135 nodes and 177,848 edges). This interactome network is then considered as the shared graph structure. There were 2,048 small molecule drug compounds and 7,793 food compounds, each one represented by a $15,135$-dim binary feature vector $\vec{x} \in \mathbb{R}^{N}$, where $x_i = 1$ if the drug or food compound targets gene $i$, and $0$ otherwise. Each drug was associated with a binary label indicating anticancer or non-anticancer class. It is worth noting that the drug dataset was highly unbalanced with only 10.2\% of anticancer drugs, and drug features were quite sparse with drugs targeting around 1.36\% of genes in average. We used the Molecular Signature Database [\citenum{Liberzon2015TheCollection}] to build  186 biological pathways (supermodules) for aggregation of genes.

\textbf{Baselines.} We compared our model against the method introduced in [\citenum{Veselkov2019HyperFoods:Foods}] (Baseline) as well as a number of state-of-the-art deep learning architectures for graph classification, i.e., Graph Convolutional Networks (GCN) [\citenum{Kipf2017}], Graph Isomorphism Networks (GIN) [\citenum{Xu2019HowNetworks}]. To allow GCNs and GINs to achieve their best possible performance, we compute graph-level outputs after each convolutional layer and combine them via concatenation, inspired by the Jumping Knowledge framework [\citenum{Xu2018RepresentationNetworks}]. Furthermore, to explore the importance of the graph augmented features with regard to these graph classification models, we trained GCNs and GINs in two settings: with raw features or graph augmented features (denoted with $+$).

 \begin{figure*}[htp!]
  \centering
  \includegraphics[scale=0.68]{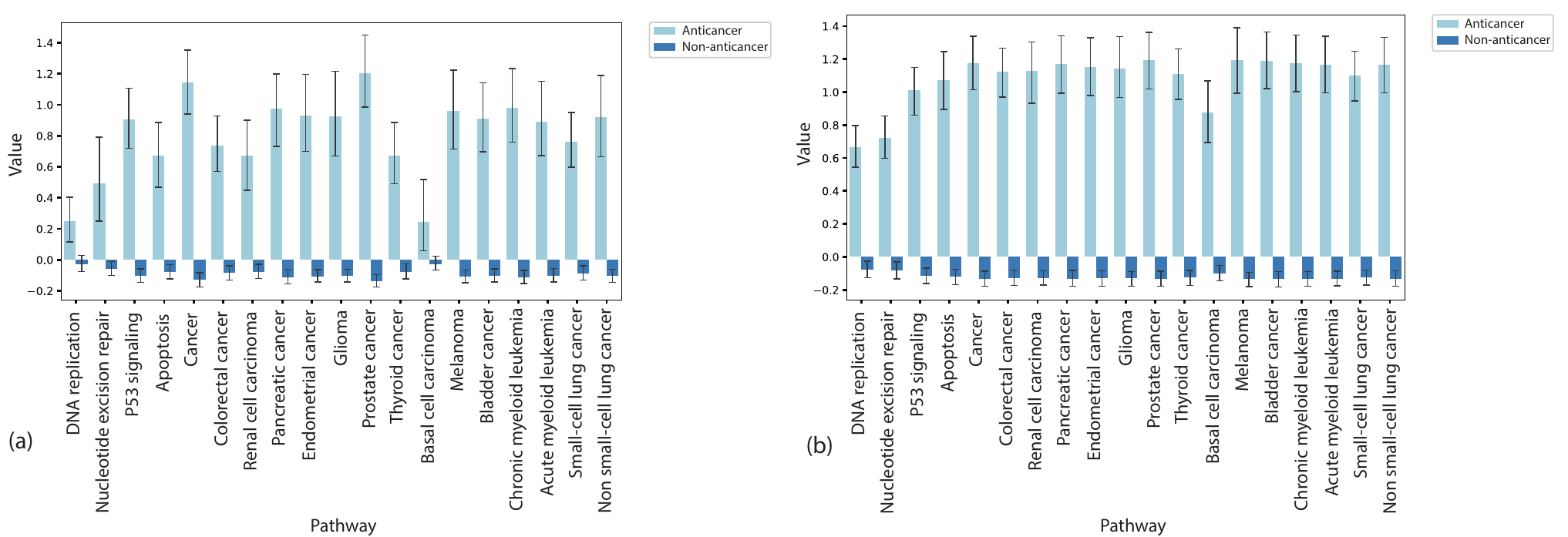}
  \caption{Visualization of drug embedding values for cancer-related pathways for (a) the baseline model, and (b) our model}
  \label{figure_barplots}
\end{figure*}

\textbf{Model configurations.} We randomly split the drug-protein dataset into train/validation/test sets in the ratio of 8:1:1 (stratified splits with respect to labels). For the configurations of GCNs and GINs, 3 GNN layers (including the input layer) are used, and 2 MLP layers follow. We apply the same graph-level mean readout for GCNs and GINs. In terms of the input graph augmented features, we use $X^{\mathcal{G}}=\left [ k_{0.1} X, k_{0.2} X, ..., k_{0.9} X \right ]$. Additionally, biological pathways are used as supermodules among genes for the SupPool layer. To balance the positve/negative classes, we re-scaled weights to be inversely proportional to class frequencies of each class during training. All models were trained on NVIDIA Tesla V100 GPUs. 

  


\begin{wraptable}{tr}{5.4 cm}
  \caption{Summary of results (\%) on anticancer drug prediction. ACC represents accuracy. F1 denotes the harmonic mean of the precision and recall. AUPR is area under the precision-recall curve. Graph-augmented features denoted with $+$.}
  
  \label{tab:results}
  \centering
  \begin{tabular}{llll}
    \toprule
    \textbf{Method} & \textbf{ACC}   & \textbf{F1} & \textbf{AUPR}  \\
    \midrule
    Baseline & 79.02 &  46.91 & 56.26     \\
    \midrule
    GCN  & 70.73 & 25.00 & 21.87 \\
    GIN   & 84.88 & 45.61 & 48.01 \\
    GCN$+$  & 69.76 & 31.11 & 27.67 \\
    GIN$+$  & \textbf{90.24} & 54.54 & 43.64  \\
    \midrule
    GAA$+$  & 88.29 & \textbf{58.62} & \textbf{62.04}  \\

    \bottomrule  
  \end{tabular}
\end{wraptable}  

\textbf{Results.} Results of our experiments are summarized in Table \ref{tab:results}. We show that our proposed model (GAA) significantly improves performance of the baseline and graph classification models in this dataset. It is worth noting that most of the metrics of GCNs and GINs using our proposed graph augmented features improve remarkably. It is also interesting to observe that embeddings of anticancer drugs show higher values in cancer-related pathways as compared to embeddings of non-anticancer drugs. Importantly, the differences are significantly increased in embeddings generated by our model (Figure \ref{figure_barplots} (b)) compared with the baseline (Figure \ref{figure_barplots} (a)). This visualization provides insights into how our model can learn biologically-meaningful embeddings that are useful for the anticancer class prediction.


\textbf{Anticancer foods prediction.}
Among the predicted anticancer compounds with a probability  $\geqslant 0.9$ we found several additional compounds to those reported in [\citenum{Veselkov2019HyperFoods:Foods}], supported by experimental evidence (see Supplementary materials), which are present in tea, root vegetables, coffee, bay, and breadfruit. Although our initial results are very encouraging, as part of our future work we are planning to do extensive sensitivity analysis of the GAA model predictions.



\section{Conclusion}
In this paper, we introduce a novel graph augmented feature for easing training of graph neural networks with sparse low-dimensional node features. Our model allows end-to-end training and prediction while providing meaningful biological embeddings. Experiments suggest that the proposed GAA model is capable of encoding both graph structure and node features in a way useful for graph classification. In this setting, our model outperforms the baseline method and several recently proposed graph neural networks by a significant margin.

\section{Acknowledgments}

The authors were supported by the ERC-Consolidator Grant No. 724228 (LEMAN) (MB, GG and SG), the Vodafone Foundation as part of the ongoing DreamLab/DRUGS project (KV, IL and GG), and the Imperial NIHR Biomedical Research Center for prospective clinical trials (MB, KV, IL and GG).

\bibliographystyle{plainnat}
\bibliography{hyperfoods_neurips}

\appendix

\newpage

\section{Supplementary Material}

%



\subsection{Food predictions}
We predicted anticancer probability of food compounds with our proposed model. Of those with a probability $\geqslant 0.9$, our model predicted 4 compounds not reported in the original baseline, with extensive experimental evidence supporting their anticancer properties (see Table \ref{table_ac_evidence}).
\renewcommand{\arraystretch}{1.5}
\begin{center}
\begin{table}[hb!]
  \caption{Predicted anticancer compounds and experimental evidence supporting anticancer properties}
\begin{tabular}{ | m{1.7cm} | m{2cm}| m{8.5cm} | }

  \hline
    \textbf{Name}     & \textbf{Anticancer probability}     & \textbf{Experimental evidence supporting anticancer properties (PMID)}   \\
    \hline
    Parthenolide & 0.96 &  11360202, 15286701, 15827332, 17556802, 17876045, 20233868, 21829151, 22109788, 23037503, 23318959, 23660068, 24065392, 24619908, 25553117, 26521947, 26824319, 27396927, 28176967    \\
    \hline
    Artonin E & 0.92 & 23225436, 23898063, 27019365, 27367662, 28356713, 28771532, 31032087\\ \hline
    Olomoucine & 0.92 & 7549905, 9436644, 9841966, 10567774, 10871858, 11679575, 11958860, 15640940, 19737069, 20010439, 24376706\\\hline
    Silibinin & 0.92 & 15476849, 18089718, 20537993, 21159609, 21954330, 22820499, 23588585, 24269256, 24440808, 25285031, 28042859, 28435252 \\

    \hline  
    
\end{tabular}
\label{table_ac_evidence}

\end{table}
\end{center}

\end{document}